\documentclass[10pt,twocolumn,letterpaper]{article}

\usepackage{cvpr}              

\usepackage{graphicx}
\usepackage{amsmath}
\usepackage{amssymb}
\usepackage{booktabs}
\usepackage{enumitem}

\usepackage[normalem]{ulem}
\usepackage{multirow}
\usepackage[pagebackref,breaklinks,colorlinks]{hyperref}

\usepackage[capitalize]{cleveref}
\crefname{section}{Sec.}{Secs.}
\Crefname{section}{Section}{Sections}
\Crefname{table}{Table}{Tables}
\crefname{table}{Tab.}{Tabs.}

\def\cvprPaperID{1806} 
\def\confName{CVPR}
\def\confYear{2022}

\begin{document}
	
	\title{Vision Transformer Slimming: Multi-Dimension Searching \\in Continuous Optimization Space} 
	
	\author{Arnav Chavan$^{*1,3}$, Zhiqiang Shen\thanks{indicates equal contribution. This work was done when Arnav was a research assistant at MBZUAI, Zhiqiang Shen is the corresponding author.}~~$^{2,3}$, Zhuang Liu$^{4}$, Zechun Liu$^5$, Kwang-Ting Cheng$^{6}$ and Eric Xing$^{2,3}$\\
		$^1$IIT Dhanbad  ~~$^2$CMU ~~$^3$MBZUAI~~$^4$UC Berkeley ~~$^5$Reality Labs, Meta Inc. ~~$^6$HKUST \\
		{\tt\small arnav.18je0156@am.iitism.ac.in  \{zhiqians,zechunl\}@andrew.cmu.edu zhuangl@berkeley.edu} \\ 
		{\tt\small timcheng@ust.hk epxing@cs.cmu.edu}
	}
	\maketitle
	
	\begin{abstract}
		This paper explores the feasibility of finding an optimal sub-model from a vision transformer and introduces a pure vision transformer slimming (ViT-Slim) framework. It can search a sub-structure from the original model end-to-end across multiple dimensions, including the input tokens, MHSA and MLP modules with state-of-the-art performance. Our method is based on a learnable and unified $\ell_1$ sparsity constraint with pre-defined factors to reflect the global importance in the continuous searching space of different dimensions. The searching process is highly efficient through a single-shot training scheme. For instance, on DeiT-S, ViT-Slim only takes $\sim$43 GPU hours for the searching process, and the searched structure is flexible with diverse dimensionalities in different modules. Then, a budget threshold is employed according to the requirements of accuracy-FLOPs trade-off on running devices, and a re-training process is performed to obtain the final model. The extensive experiments show that our ViT-Slim can compress up to 40\% of parameters and 40\% FLOPs on various vision transformers while \emph{increasing} the accuracy by $\sim$0.6\% on ImageNet-1K. We also demonstrate the advantage of our searched models on several downstream datasets. Our code is available at \url{https://github.com/Arnav0400/ViT-Slim}.
	\end{abstract}
	
	\section{Introduction}
	\label{sec:intro}
	
	Transformer~\cite{vaswani2017attention} has been a strong network model for various vision tasks, such as image classification~\cite{dosovitskiy2020image,touvron2021training,liu2021swin,wu2020visual,shen2021sliced}, object detection~\cite{carion2020end,zhu2020deformable,sun2021rethinking}, segmentation~\cite{zheng2021rethinking,xie2021segformer,wang2021end}, etc. It is primarily composed of three underlying modules: Multi-Head Self-Attention (MHSA), Multi-Layer Perceptron (MLP) and Image Patching Mechanism. The major limitation of ViT in practice is the enormous model sizes and excessive training and inference costs, which hinders its broader usage in real-world applications. Thus, a large body of recent studies has focused on compressing the vision transformer network through searching a stronger and more efficient architecture~\cite{AutoFormer,chen2021glit,chen2021chasing,rao2021dynamicvit,su2021vision} for better deployment on various hardware devices.
	
	However, many commonly-used searching strategies are recourse-consuming, such as the popular reinforcement learning and evolutionary searching approaches. Single-path one-shot (SPOS) \cite{guo2020single} is an efficient searching strategy and promising solution for this task, while it still needs to train the {\em supernet} for hundreds of epochs and then evaluate thousands of {\em subnets} for finding out the optimal sub-architecture, which is still time-consuming, typically with more than tens of GPU days.
	
	Recently, there are some works utilizing Batch Normalization (BN) scaling parameter as the indicator for the importance of operations to prune or search subnets, such as Network Slimming~\cite{liu2017learning}, SCP~\cite{kang2020operation}, BN-NAS~\cite{chen2021bn}, etc., since BN's parameter is an existing factor and light-weight measure metric in the network for the importance of subnets. This searching method can offer 10$\times$ speedup of training than SPOS in general. But in practice, not all networks contain BN layers, such as the transformers. Also, there are many unique properties in a transformer design like the dependency of input tokens from shallow layers to deep layers. Simply utilizing such a strategy is not necessarily practical or optimal for the newer transformer models. 
	
	\begin{table*}[]
\centering
\resizebox{\textwidth}{!}{
\begin{tabular}{l|cccccc}
\hline
{Method} & {Target Arch}      & {Searching Space}        & {Searching Method}         & {Searching Time}   &    Inherit Pre-train   &    Reduce Params \& FLOPs          \\ \hline
GLiT \cite{chen2021glit} & SA + 1D-CNN & Discrete, Pre-defined & Two stage evolutionary & 200-ep  & No & Both \\
S$^2$ViTE  \cite{chen2021chasing}  & ViT/DeiT family & Continous, limited & Iterative Prune \& Grow & 510 g-hrs (600-ep) & Yes & Both    \\
Dynamic-ViT \cite{rao2021dynamicvit} & ViT/DeiT family & Dynamic Patch Sel. & Layerwise Prediction module & 26 g-hrs (30-ep) & Yes & FLOPs only \\
Patch-Slimming \cite{tang2021patch} &   ViT/DeiT family  &  Layerwise  Patch  Sel.  &  Top-Down layerwise  & 31 g-hrs (36-ep) & Yes  &  FLOPs only \\
ViTAS \cite{su2021vision}  & Customized ViT   & Discrete, Pre-defined & Evolutionary  & 300-ep &  No & Both \\
AutoFormer \cite{AutoFormer}   & Customized ViT  & Discrete, Pre-defined  & Evolutionary   & 500-ep  & No & Both \\
ViT-Slim (Ours) & ViT/DeiT/Swin, etc.    & \bf Continuous, all modules  & \bf One-shot w/ $\ell_1$-sparsity  & \bf 43 g-hrs (50-ep) &\bf Yes  & \bf Both \\ \hline
\end{tabular}
}
\vspace{-0.12in}
\caption{Feature-by-feature comparison of compression and search approaches for vision transformers. ``g-hrs'' indicates GPU hours.}
\label{tab:comparemethod}
\vspace{-0.12in}
\end{table*}
	
	Consequently, the main remaining problem is that there is no BN layer involved in conventional transformer architectures, so we cannot directly employ the scaling coefficient in BN as the indicator for searching. To address this, in this work we propose to incorporate the explicitly soft masks to indicate the global importance of dimensions across different modules of a transformer. We consider jointly searching on all three dimensions in a transformer end-to-end, including: layerwise tokens/patches, MHSA and MLP dimensions. In particular, we design additional differentiable soft masks on different modules for the individual dimension of a transformer, and the $\ell_1$-sparsity is also imposed to force the masks to be sparse during searching. We only need a few epochs to finetune these mask parameters (they are initialized to 1 so as to give equal importance to all dimensions at the beginning of search) together with the original model's parameters for completing the whole searching process, which is extremely efficient. For the token search part, we apply a {\em tanh} over masks to avoid exploding mask values which is observed empirically.
	
	We call our method {\em ViT-Slim}, a joint sparse-mask based searching method with an implicit weight sharing mechanism for searching a better sub-transformer network. This is a more general and flexible design than previous BN-based approaches since we have no requirement of BN layers in the networks. This is more friendly to transformers and a feature-by-feature comparison with other ViT compression methods is shown in Table~\ref{tab:comparemethod}. One core advantage of our method is the efficiency of searching, we can inherit the pre-trained parameters and conduct a fast search upon it.  Another advantage is the zero-cost subnet selection and high flexibility. In contrast to the SPOS searching that requires to evaluate thousands of subnets on validation data, once we complete the searching process, we can obtain countless subnetworks and the final structure can be determined by the requirement of accuracy-FLOPs trade-off of the real devices that we deploy our model on, without any extra evaluation. The last advantage is that we can search for a finer-grained architecture such as the different dimensionalities in different self-attention heads, as our search space is continuous in them. This characteristic allows us to find the architecture with unique individual dimensions and shapes in different layers and modules, which would always find out a better subnet than other counterparts.
	
	Comprehensive experiments and ablation studies are conducted on ImageNet-1K~\cite{deng2009imagenet}, which show that ViT-Slim can compress  up  to  40\%  of  parameters  and  40\%  FLOPs  on  various vision transformers like DeiT~\cite{pmlr-v139-touvron21a}, Swin~\cite{liu2021swin} without any compromising accuracy (in some circumstances our compressed model is even better than the original one). We also demonstrate the advantage of our searched models on several downstream datasets.
	
	Our main contributions are: 
	\vspace{-0.08in}
	\begin{itemize}[leftmargin=0.15in]
		\addtolength{\itemsep}{-0.1in}
		\item We introduce ViT-Slim, a framework that can jointly perform an efficient architecture search over all three modules - MHSA, MLP and Patching Mechanism in vision transformers. We stress that our method searches for structured architectures which can bring practical efficiency on modern hardware (e.g., GPUs).
		\item We empirically explore various structured slimming strategies by sharing weights in candidate structures, and provide the best performing structure by employing a continuous search space in contrast to a pre-defined discrete search space in existing works.
		\item Our method can perform directly over pre-trained transformers by employing a single shot searching mechanism for all possible budgets, eliminating the need of search-specific pre-training of large models and performing repeated searching for different modules/budgets.  
		\item We achieve state-of-the-art performance at different budgets on ImageNet-1K across a variety of ViT compression and search variants. Our proposed ViT-Slim can compress up to 40\% of parameters and 40\% FLOPs while increasing  the  accuracy by $\sim$0.6\%.
	\end{itemize}
	
	\vspace{-0.15in}
	\section{Related Work}
	\vspace{-0.04in}
	
	\noindent{\textbf{Efficient Models and Architecture Search}}
	Neural network compression has been recognized as an important technology for applying deep neural network models to equipment with limited resources. The compression research extends from channel pruning~\cite{he2017channel,liu2017learning,ye2018rethinking}, quantization and binarization~\cite{liu2020reactnet,bulat2019xnor,hubara2017quantized,zhou2016dorefa,choi2018pact,phan2020binarizing}, knowledge distillation~\cite{hinton2015distilling,romero2014fitnets,shen2020label,muller2019does,shen2020meal,shen2021afast}, compact neural network design~\cite{ma2018shufflenet,zhang2018shufflenet,wu2018shift,chollet2017xception,iandola2016squeezenet} to architecture search~\cite{real2019regularized,zoph2018learning,zoph2016neural}.
	
	Specifically, MobileNets~\cite{howard2017mobilenets,sandler2018mobilenetv2} proposed to decompose the convolution filters to depth-wise convolution and point-wise convolution for reducing the parameters in convolutional neural networks (CNNs). EfficientNet~\cite{tan2019efficientnet} proposed to search the uniform scaling ratio among all dimensions of depth/width/resolution to achieve much better accuracy and efficiency. Network Slimming~\cite{liu2017learning} used the BN parameters as the scaling factors to find the best sub-structure. JointPruning~\cite{liu2021joint} jointly searched the layer-wise channel number choices together with the depth and resolution for finer-grained compression. NetAdapt~\cite{yang2018netadapt} and AMC~\cite{He2018AMCAF} adopted the feedback loop or the reinforcement learning method to search the channel numbers for the CNNs. Besides, many neural architecture search (NAS) methods are targeting at exploring the operation choices (e.g., $3\times3$, $5\times5$, $7\times7$ convolutions) for architectures. For instance, SPOS~\cite{guo2020single} built a supernet that contained all the possible choices and used the evolutionary search for a subnetwork in the supernet. DARTS~\cite{liu2018darts}, FB-Net~\cite{wu2019fbnet} and ProxylessNAS~\cite{cai2018proxylessnas} used gradient-based method to update the mask associating with each operation choices. However, these NAS methods defined on the discrete operation search spaces can hardly be generalized to tackle the problem of continuous channel number search.

	\noindent{\textbf{Efficient Vision Transformers.}} There are several works exploring on this direction~\cite{chen2021chasing,rao2021dynamicvit,su2021vision,chen2021glit,AutoFormer}. 
	Patch-Slimming \cite{tang2021patch} explored the direction to improve the efficiency of transformers by sequentially pruning patches from top-to-bottom layers. 
	Similarly, Dynamic-ViT \cite{rao2021dynamicvit} explored dynamic patch selection based on the input patches. 
	They employed multiple hierarchical prediction modules to estimate the importance score of each patch. 
	However, patch pruning did not improve parameter efficiency. 
	ViTAS \cite{su2021vision} used the evolutionary algorithm to search optimal architecture at a target budget. 
	However, their search space was discrete, pre-defined and thus limited. 
	GLiT \cite{chen2021glit} introduced a locality module to model local features along with the global features. But their method used CNNs along with the attention and performed an evolutionary search over global and local modules. BigNAS \cite{yu2020bignas} introduced the single-stage method to generate highly efficient child models by slicing weight matrices. Based on this, AutoFormer \cite{AutoFormer} showed that weight entanglement was a better alternative than defining weight matrices for every possible sub module in architecture search and using an evolutionary algorithm to search for optimal sub-networks. But all of them had a limited discrete search space as \cite{su2021vision} due to the adaptation of an evolutionary algorithm for searching.
	S$^2$ViTE  \cite{chen2021chasing} provided an end-to-end sparsity exploration for vision transformers with an iterative pruning and growing strategy. Their structured pruning method eliminated complete attention heads and MLP neurons based on a scoring function calculated using the Taylor expansion for the loss function and $\ell_{1}$-norm, respectively. We argue that eliminating complete attention heads is a sub-optimal choice and limits the learning dynamics of a transformer. Allowing the model to determine optimal dimensions for every attention head (instead of eliminating complete heads) is a better alternative for pruning MHSA module. 
	
	\begin{figure}[t]
		\centering
		\includegraphics[width=0.485\textwidth]{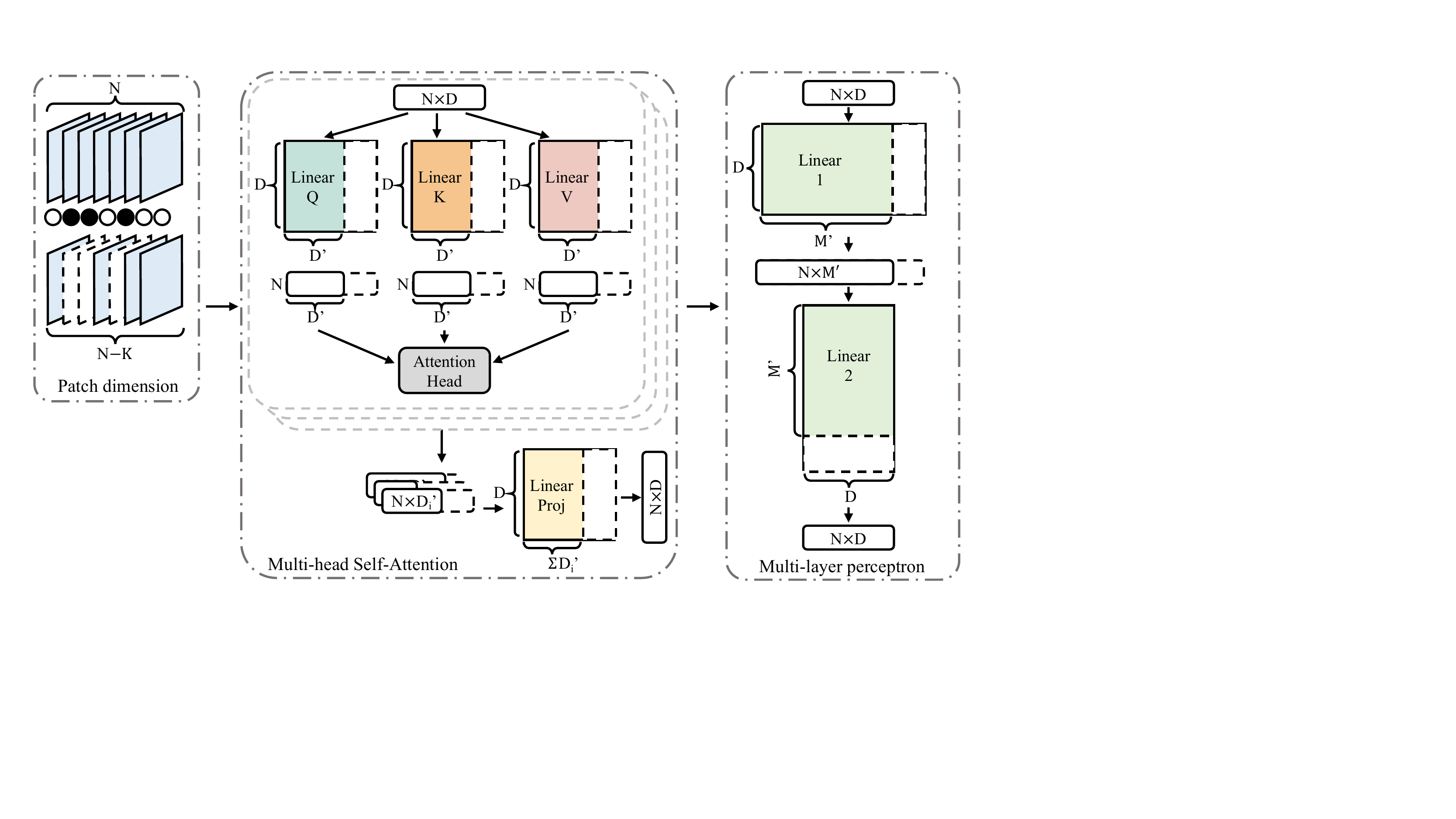}
		\vspace{-0.23in}
		\caption{An overview of our proposed ViT-Slim framework. D' and M' are the dimensions after slimming operation.}
		\label{fig:overview}
		\vspace{-0.05in}
	\end{figure}
	
	\section{Proposed Approach}
	
	\subsection{Overview and Motivation}
	In this section, we start by working around some important questions raised by existing works, viz:
	\vspace{-0.1in}
	\begin{itemize}[leftmargin=0.15in]
		\addtolength{\itemsep}{-0.08in}
		\item Can we take advantage of the existence of only fully connected layers in transformers and make the search space continuous and hence much larger than existing works without larger memory or compute overhead?
		\item What is the optimal structure configuration that has to be searched in ViT family, and can joint search of architecture (MHSA/MLP dimension search) and layer-wise data flow mechanism inside the architecture (layer-wise patch selection) be coupled together in a single shot setup?
		\item What is the impact of individual modules to the final model's performance and can vision transformers which are originally designed to have an isotropic structure benefit from a highly non-uniform structure?
	\end{itemize}
	\vspace{-0.1in}
	
	An overview of our framework is shown in Figure~\ref{fig:overview}. In the following, we will discuss (i) How to achieve a continuous search space; (ii) Identify optimal search space; and (iii) Single-shot architecture search with the $\ell_1$-sparsity.
	
	\subsection{Achieving A Continuous Search Space}
	
	One-shot NAS methods for CNNs \cite{guo2020single, chu2021fairnas, tan2019efficientnet} explicitly define multiple decoupled candidate blocks for every layer for training a central supernet. This strategy is suitable for CNNs as the candidate blocks at every layer come from a wide variety of sub-architectures so as to maintain the property of sampling diverse subnets from the supernet while searching. This is not true for transformers, as they are internally composed of multiple fully connected layers stacked in different configurations for different modules. The fact that the core blocks are all composed of fully connected layers opens the possibility to expand the search space by sharing weights between the candidate fully connected layers inside any block. 
	
	Consider a fully connected layer which has to be searched for optimal output dimension given an input dimension $D_{in}$, conventional method is to define multiple candidate layers with output dimensions from pre-defined search space and search the optimal layer from them with a suitable searching algorithm. However, there are a couple of disadvantages: 1) Longer search time as every candidate layer needs to be at least partially trained for searching; 2) Larger memory footprint contributed by weight matrices of every candidate layer. We propose to solve these issues by weight sharing between all possible candidate layers. We fix the maximum permissible output dimension to $D_{max}$ and define a super weight matrix $W_{sup} \in D_{in} \times D_{max}$. The candidate layer weights can be easily sliced from $W_{sup}$.
	
	To achieve a continuous search space, we adopt a single-stage method to rank the importance of every dimension of supernet weights, taking inspiration from \cite{liu2017learning} we use $\ell_1$-sparsity to achieve it. We first pre-train the supernet until convergence (in practice, our method would work directly over pre-trained networks eliminating the need of supernet training). We then define masks corresponding to every dimension that has to be searched. The magnitude of mask values correspond to the importance score of respective dimension and thus we initialize all of them by 1. Once the pretrained weights are loaded we induce these masks into the model. Considering the previous example of a single fully-connected layer where the output dimension is to be searched of a layer with weight matrix $W_{sup}$, a mask $z \in D_{max}$ is defined. A dot product between $W_{sup}$ and $z$ gives the candidate weight matrix to be used in forwarding propagation. The search algorithm employs a loss function which is a combination of pre-training loss function (Cross-Entropy in classification tasks) and $\ell_1$-norm on masks. This combined loss drives the mask values towards 0 while minimizing the target loss function. In a way the optimization landscape implicitly drives the masks to rank themselves according to their impact on final performance.
	
	\subsection{Identifying Optimal Search Space}
	The fundamental question in NAS is defining a search space. This section presents our defined search space.
	
	\noindent{\textbf{Flexible MHSA and MLP Modules.}} 
	Recent works have followed two methodologies for defining the search space for MHSA module: 1) Searching for the number of heads in every distinct MHSA module \cite{LinLYHR20, chen2021chasing}; and/or 2) Searching for a common feature dimension size from a pre-defined discrete sample space for all attention heads in any particular MHSA module \cite{chen2021glit, su2021vision, AutoFormer}. These methods have shown some solid results but they are not completely flexible. A greater degree of flexibility can be achieved if every attention head can have a distinct feature dimension size. Assuming a super-transformer network with a maximum of $L$ MHSA modules each with a maximum permissible number of heads set to $H$. This gives a total of $L\times H$ unique attention heads. If we fix the maximum permissible feature dimension size to $d$, the size of the equivalent search space is equal to $(d+1)^{L\times H}$. Searching in such a massive search space is computationally very difficult. However, such a diverse search space has a couple of advantages: 1) The search algorithm can be more flexible in adapting the super-transformer to smaller architectures while maintaining the performance; and 2) The extracted architectures would be much more efficient as the search algorithm can push the feature dimensions of least important attention heads to even zero, decreasing the FLOPs substantially. We use the complete $(d+1)^{L\times H}$ search space in our method.
	
	Similarly, for the MLP, there are a total of $L$ modules throughout the network with existing works using discrete and limited search space to search the optimal feature dimension. If we fix the maximum permissible feature dimension size to $M$, the search space which can be explored by our method is equal to $(M+1)^{L}$. Combining MHSA and MLP together in a single search mechanism is quite straightforward. Thus, it will generate a much more diverse set of child networks compared to all the existing works.
	
	\noindent{\textbf{Patch Selection.}} 
	Patch slimming \cite{tang2021patch} showed that MHSA aggregates patches and consequently cosine-similarity between the patches increases exponentially layer-wise becoming as large as 0.9 in the final layers. This opens the possibility to eliminate a large number of deeper patches and a few unimportant shallow patches too. Inducing patch-selection with MHSA and MLP search can extract more efficient architectures with reduced FLOPs at the same parameter count. While intuitively, a dynamic way is more promising on the patch dimension as the selected patches should be aligned with the input images to reflect the importance of different regions in the image.
	
	\subsection{Single Shot Arch. Search with $\ell_1$-Sparsity}
	\label{search}
	The main objective of our search method is to rank the mask values according to their impact on final performance. Once ranked, we eliminate the dimensions having the least mask values. Let $f:\mathbb{R}^x \rightarrow \mathbb{R}^y$ denote a vision transformer network which learns to transform input $\mathbf{x}$ into target outputs $\mathbf{y}$ with weights $\mathbf{W}$ and intermediate activations/tensors $\mathbf{T} \in \mathbb{R}^d$ constructed from weights $\mathbf{W}$ and input $\mathbf{x}$. We define $\mathbf{z} \in \mathbb{R}^d$ as a set of sparsity masks, where $z_i \in \mathbf{z}$ refers to the mask associated with the intermediate tensor $\mathbf{t}_i \in \mathbf{T}$. To apply the mask, $z_i$ is multiplied with corresponding entries of $\mathbf{t}_i$. The optimization objective can be stated as:
	\begin{equation}
		\min_{\mathbf{W}, \mathbf{z}} \mathcal{L_{CE}}(f(\mathbf{z}\odot \mathbf{T(W, \mathbf{x})}), \mathbf{y}) + \lVert \mathbf{z} \rVert_{1}, 
		\label{eq_opt_problem}
	\end{equation}
	We introduce uniform masking to search optimal dimension size for each distinct head in MHSA modules, dimension size for each MLP module and layerwise most important patches. Consider a transformer network with $L$ layers of MHSA+MLP blocks and each block consisting of $H$ self-attention heads. The input tensor to each MHSA layer $\mathbf{t}_{a_{l}} \in \mathbb{R}^{N \times D}$ where $N$ is number of patches and $D$ is global feature dimension size. Inside every head $i$ of MHSA module, $\mathbf{t}_{a_{l}}$ is transformed with fully connected layers to $\mathbf{q_{i}} \in \mathbb{R}^{N \times d}$, $\mathbf{k_{i}} \in \mathbb{R}^{N \times d}$ and $\mathbf{v_{i}} \in \mathbb{R}^{N \times d}$, $d$ denoting feature dimension of each self-attention head. We define mask $\mathbf{z_a} \in \mathbb{R}^{L \times H}$, and corresponding vectors in $\mathbf{z_a}$, $z_{a_{l,h}}\in\mathbb{R}^{d}$ corresponding to $l^{th}$ layer and $h^{th}$ head. The total possibilities for MHSA modules across the network that can be explored by our method is thus $(d+1)^{L\times H}$. The computations inside MHSA module with sparsity masks is:
	\begin{equation}
		A_{i} = softmax((q_{i} \odot z_{a_{l,h}}) \times (k_{i} \odot z_{a_{l,h}})^{T} / \sqrt{d}) 
	\end{equation}
	\begin{equation}
		O_{i} = A_{i} \times (v_{i} \odot z_{a_{l,h}}) \\
	\end{equation}
	\begin{equation}
		\mathbf{t_{m_{l}}} = projection([O_{1}, O_{1}, ..., O_{H}])
	\end{equation}
	where $\mathbf{t_{m_{l}}} \in \mathbb{R}^{N \times D}$ is the output from MHSA block which in turn becomes the input to MLP block. Inside MLP block, $\mathbf{t_{m_{l}}}$ is projected to a higher-dimensional space through a fully connected layer $f_{1}$ to form an intermediate tensor $\mathbf{t}_{e_{l}} \in \mathbb{R}^{N \times M}$ which is again projected back to ${R}^{N \times D}$ through another fully connected layer $f_{2}$. We define mask $\mathbf{z_m} \in \mathbb{R}^{L}$, and corresponding vectors in $\mathbf{z_m}$, $z_{m_{l}}\in\mathbb{R}^{M}$ corresponding to $l^{th}$ layer. The total possibilities for MLP modules across the network is thus $(M+1)^L$ by our method. The following computation shows the interaction of masks with MLP module:
	\begin{equation}
		\mathbf{{t}_{e_{l}}} = f_{1}(\mathbf{t_{m_{l}}}) \odot z_{m_{l}}, \mathbf{t}_{a_{l+1}} = f_{2}(\mathbf{t_{e_{l}}})
	\end{equation}

	\noindent{\textbf{Solving Patch Dependency Across Layers in Patch Dim.}}
	For patch selection, we define a distinct mask value corresponding to each patch in every layer and eliminate the patches corresponding to a lower mask value. A problem arises due to the global single shot search that there may be anomalous instances where the same patch is eliminated in a shallower layer but exits in the deeper layer. In practice, such anomalous instances are limited indicating that $\ell_1$-Sparsity based search strategy aligns the patches as per their importance and a shallow eliminated patch implicitly forces it's deeper counterpart to have less importance and consequently eliminates it. To counter these limited anomalous patches, once a patch is eliminated from an earlier layer, we eliminate it from further layers too, while imposing budget. Additionally, we apply a $tanh$\footnote{initialized to 3.0 which is equivalent to $\sim$1.0 after {\em tanh}.} activation function over patch-specific masks before taking a dot product with the patches to stop mask values from exploding.
	
	\vspace{-0.1in}
	\subsubsection{Searching Time Analysis}
	\vspace{-0.06in}
	Our method directly works over pre-trained models, eliminating the need of training a search-specific model. We induce sparsity masks on a pre-trained model and jointly optimize the masks and model weights with a combination of CE loss and $\ell_{1}$-norm of masks. In our setup, we fix the search epochs to 50 for all searches. This translates to $\sim$43 GPU hrs for DeiT-S and $\sim$71 GPU hrs for DeiT-B. At the end of search, masks are ranked according to their values.
	
	\vspace{-0.1in}
	\subsubsection{Retraining with Implicit Budget Imposition}
	\vspace{-0.06in}
	Once ranked, depending upon the target budget, low-rank dimensions/patches are eliminated from the network. For MHSA+MLP joint search, budget approximately translates to FLOPs and Params budget too due to the linear relation between the number of dimensions and FLOPs/Params, and inducing patch selection further decreases FLOPs. Once the budget-specific structure is extracted, it is retrained with exactly the same setup as the pre-trained model. This allows the weights to adjust from continuous masks in searching space to binary/non-existing masks in the final structure.  
	
	\section{Experiments}
	
	In this section, we first explore the contribution of each individual component in the final performance of the ViT~\cite{dosovitskiy2020image}/DeiT~\cite{touvron2021training} model and search for the optimal unidimensional searched model. We then move to joint search combining all three components and show that our method outperforms all existing architecture search and pruning methods. We also show the applicability of our method to other transformer architectures, such as Swin~\cite{liu2021swin}. Finally, we further show the performance of searched models on the downstream datasets in a transfer learning setup.
	
	\vspace{-0.04in}
	\subsection{Training Procedures and Settings }
	\vspace{-0.02in}
	
	There are three steps in our workflow for the ViT-Slim framework, including:
	
	\noindent{\textbf {(i) One-shot Searching.} We use pre-trained weights to initialize existing vision transformer models and use them as our supernet. We then induce sparsity masks into the model depending upon the dimension to search and jointly train the weights and masks with the loss function given in Equation~\ref{eq_opt_problem} for 50 epochs with a constant learning rate of 5e-04 and AdamW optimizer with a weight decay of 1e-03. The batch size is set to 1,024 for DeiT-S and 512 for DeiT-B. We also employ stochastic depth \cite{huang2016deep}, cutmix \cite{yun2019cutmix}, mixup \cite{zhang2017mixup}, randaugment \cite{cubuk2020randaugment}, random erasing \cite{zhong2020random}, etc., as augmentations following DeiT while searching.}
	
	\noindent{\textbf {(ii) Budget Selection.} Once the search is complete, the masks are ranked according to their values after the searching step. Depending upon the target budget of compression, low-rank dimensions are eliminated from the supernet to extract the final structure of searched model.}
	
	\noindent{\textbf {(iii) Re-training} Finally, we retrain the extracted compressed structure for 300 epochs with the same setting on which it was originally pre-trained in \cite{touvron2021training,liu2021swin}}.
	
	\vspace{-0.04in}
	\subsection{Unidimensional Search}
	\vspace{-0.02in}
	
	To show the impact of MHSA and MLP modules independently in the final model's performance, we search for optimal dimensions in both of them separately. We induce sparsity masks in respective modules of DeiT-S and search two supernets each with a sparsity weight of 1e-04 for MLP and MHSA dimensions. The total number of masks induced in MLP is $4 \times$ that of MHSA due to the existence of $4 \times$ more dimensions in MLP module. The post-search accuracy and final accuracy at different budgets are shown in Table \ref{tab:uni}. At higher budgets, compressed models even perform better than pre-trained model giving as much as 1\% boost in accuracy. MHSA search performs better than MLP search at the same budgets, but MLP achieves a better degree of parameter and FLOPs compression. MHSA at 40\% and MLP at 60\% have the same FLOPs but MLP outperforms MHSA. Similarly, MHSA at 40\% and MLP at 70\% having the same number of parameters, MLP outperforms MHSA by a fair margin of 1.2\%. This clearly shows that it is much easier to compress MLP dimensions as compared to MHSA dimensions to achieve the same target FLOPs/Parameters, indicating the importance of MHSA to be greater than MLP.
	
	\begin{table}[]
\centering
\resizebox{0.48\textwidth}{!}{
\setlength\tabcolsep{1.6pt}
\begin{tabular}{cc|cc|cc}
Budget (\%) & Module                & \#Params (M) & FLOPs (B) & Top-1 (\%) & Top-5 (\%) \\ \midrule
100         & -                     & 22.0         & 4.6       & 79.90      & 95.01      \\ \midrule
Post-Search & \multirow{5}{*}{MHSA} & 22.0         & 4.6       & 76.85      & 93.70      \\
70          &                       & 19.9         & 4.1       & 80.90      & 95.44      \\
60          &                       & 19.2         & 3.9       & 80.63      & 95.31      \\
50          &                       & 18.5         & 3.7       & 80.10      & 95.07      \\
40          &                       & 17.8         & 3.5       & 79.61      & 94.73      \\ \midrule
Post-Search & \multirow{5}{*}{MLP}  & 22.0         & 4.6       & 76.85      & 93.70      \\
70          &                       & 17.8         & 3.8       & 80.80      & 95.37      \\
60          &                       & 16.4         & 3.5       & 80.39      & 95.28      \\
50          &                       & 15.0         & 3.2       & 79.89      & 95.05      \\
40          &                       & 13.5         & 2.9       & 79.20      & 94.75     
\end{tabular}
}
\vspace{-0.1in}
\caption{Performance of {\bf  DeiT-S}~\cite{pmlr-v139-touvron21a} for MHSA and MLP dimension search. Budget indicates the \% of active dimensions of respective search modules across the network.}
\label{tab:uni}
\end{table}
	
	\vspace{-0.04in}
	\subsection{ Partial Combination for Parameter Search}
	\vspace{-0.02in}
	\begin{table}[]
\vspace{-0.12in}
\centering
\resizebox{0.48\textwidth}{!}{
\begin{tabular}{cc|cc|cc}
\multirow{2}{*}{$W_{1}$} & \multirow{2}{*}{$W_{2}$} & \multicolumn{2} {c|} {Post Search Accuracy} & \multicolumn{2} {c} {Final Accuracy }                \\ \cline{3-6} 
   &   &  Top-1 (\%) &  Top-5 (\%) & Top-1 (\%)                    & Top-5 (\%)                    \\ \midrule
1e-04             & 1e-04            & 76.54             & 93.37             & 79.10                         & 94.65                         \\
3e-04             & 1e-04            & 76.34             & 93.36             & 79.00                           & 94.70 \\
4e-04             & 1e-04            & 76.40             & 93.39             & 78.71 & 94.57 \\
2e-04             & 4e-05            &  \textbf{\uline{76.69}}    & 93.61    & 79.17                & 94.72                \\
2e-04             & 5e-05            & 76.68             & \textbf{\uline{93.64}}             & \textbf{\uline{79.20}}                & \textbf{\uline{94.75}}               
\end{tabular}
}
\vspace{-0.1in}
\caption{$W_{1}$ (MHSA sparsity weight) and $W_{2}$ (MLP sparsity weight) grid search for {\bf DeiT-S}~\cite{pmlr-v139-touvron21a} ({\bf ViT-Slim$_{\bf PS}$}). {\em Post search} indicates post search soft accuracy of supernet. {\em Final accuracy} indicates final accuracy of the compressed models at 60\% budget.}
\label{tab:grid1}
\vspace{-0.12in}
\end{table}
	Next, we combine MHSA and MLP in a single supernet search. The most important hyperparameter to control is the sparsity weight for each of different modules. Based on the fact that MLP is easy to compress and has $4 \times$ more dimensions than MHSA, we expect the optimal sparsity weight to be in a similar ratio. We start with an equal sparsity weight of 1e-04 and do a thorough grid search to achieve the optimal performance as shown in Table \ref{tab:grid1}. We search and retrain at 60\% budget for a fair comparison. From the results, it is clear that post-search accuracy directly reflects the final model accuracy. As expected, the optimal sparsity weights 2e-04 and 5e-05 are in the ratio of 4:1.
	
	\begin{table}[]
\centering
\resizebox{0.48\textwidth}{!}{
\setlength\tabcolsep{2pt}
\begin{tabular}{@{}cc|cc|cc@{}}
B-MLP & B-MHSA & \#Params (M) & FLOPs (B) &  Top-1 (\%) &  Top-5 (\%) \\ \midrule
100        & 100         & 22.0       & 4.6       & 79.90         & 95.01          \\
80         & 80          & 17.7       & 3.7       & 80.60         & 95.29          \\
70         & 70          & 15.6       & 3.3       & 80.03         & 95.05          \\
60         & 60          & 13.5       & 2.8       & 79.20         & 94.75          \\
50         & 50          & 11.4       & 2.3       & 77.94         & 94.14         
\end{tabular}
}
\vspace{-0.1in}
\caption{Performance of {\bf DeiT-S}~\cite{pmlr-v139-touvron21a} ({\bf ViT-Slim$_{\bf PS}$}) for MLP and MHSA joint dimension search. Budget indicates the \% of active MHSA and MLP dimensions across the network respectively.}
\label{tab:mhsamlp}
\vspace{-0.05in}
\end{table}
	We retrain the optimal searched model at multiple budgets as shown in Table \ref{tab:mhsamlp}. The performance of the model is intact and even better than the pre-trained model up to a budget of 70\% after which the accuracy starts to drop. This translates to better performance with 30\% FLOPs and parameter reduction. However, going from 60\% to 50\% budget shows a drastic drop in performance. We name these budgeted model family as ViT-Slim$_{\bf{PS}}$ indicating direct parameter search or partial search without patch selection.

	\noindent{\textbf{Discussion: Budgeted FLOPs and Parameter Reduction}}
	The budget indicates the number of active dimensions that exist in the final searched model for the respective modules. However, for MHSA and MLP joint search, budget translates to final FLOPs and parameter budget due to a linear relation between them. This can be helpful to deploy large transformer networks as per the target FLOPs/Parameter budget on specific hardware.
	
	\vspace{-0.04in}
	\subsection{Multidimensional Joint Search}
	\vspace{-0.02in}
	\begin{table}[]
\vspace{-0.09in}
\centering
\resizebox{0.37\textwidth}{!}{
\begin{tabular}{ccc|cc}
\multirow{2}{*}{$W_{1}$} & \multirow{2}{*}{$W_{2}$} & \multirow{2}{*}{$W_{3}$} & \multicolumn{2} {c} {Post Search Accuracy}               \\ \cline{4-5}
   & &   &  Top-1 (\%) &  Top-5 (\%)                    \\ \midrule
2e-04             & 5e-05 & 1e-04            &  \textbf{\uline{76.92}}            & 93.54       \\
2e-04             & 5e-05 & 2e-04            &                         76.76             & 93.54    \\
2e-04             & 5e-05 & 5e-05            &                         76.79             &  \textbf{\uline{93.55}}        \\  
\end{tabular}
}
\vspace{-0.1in}
\caption{$W_{1}$ (MHSA sparsity weight), $W_{2}$ (MLP sparsity weight) and $W_{3}$ (Patch sparsity weight) grid search for {\bf DeiT-S} ({\bf ViT-Slim$_{\bf JS}$}).}
\label{tab:grid2}
\vspace{-0.02in}
\end{table}
	\begin{table}[]
\vspace{-0.06in}
\centering
\resizebox{0.48\textwidth}{!}{
\begin{tabular}{c|cc|cc}
\multicolumn{1}{l|}{Budget} & \#Params (M) & FLOPs (B) & Top-1 (\%) & Top-5 (\%) \\ \midrule
100                                    & 15.6       & 3.3       & 80.03      & 95.01      \\
80                                     & 15.6       & 3.1       & 79.91      & 94.97      \\
70                                     & 15.6       & 2.9       & 79.72      & 94.91      \\
60                                     & 15.6       & 2.8       & 79.51      & 94.75  \end{tabular}
}
\vspace{-0.1in}
\caption{Performance of {\bf DeiT-S}~\cite{pmlr-v139-touvron21a} ({\bf ViT-Slim$_{\bf JS}$}) for MLP, MHSA and patch selection joint search. Budget indicates the \% of {\bf active patches} across the network. Budget for both MHSA and MLP is fixed at 70\% across models.}
\label{tab:patch}
\vspace{-0.12in}
\end{table}
	Finally, a joint search over all three dimensions - MHSA, MLP and Patch Selection is performed. Once we identify the optimal MHSA and MLP sparsity weights as 2e-04 and 5e-05, respectively, we do a grid search over patch sparsity as shown in Table \ref{tab:grid2}. Optimal patch sparsity weight used in all our experiments is 1e-04.
	To show the effect of patch selection at different budgets, we fix the MHSA and MLP budget at 70\% (as 70\% keeps the performance intact as shown in Table \ref{tab:mhsamlp}) and retrain at multiple patch-selection budgets as shown in Table \ref{tab:patch}. Eliminating as much as 40\% patches doesn't cause massive degradation in performance. At 80\% budget, performance matches that of pre-trained DeiT-S (79.9\% Top-1). We name this model family with multidimensional search as ViT-Slim$_{\bf{JS}}$ indicating joint search.
	
	\vspace{-0.04in}
	\subsection{Other Architectures}
	\vspace{-0.02in}
	We further show the efficacy of our method on DeiT-B (ViT-Slim-B$_{\bf{PS}}$) with the same set of hyperparameters. We show a thorough comparison with the existing method in Table \ref{tab:baseswin}. Our model's budget is set to 60\% which is equivalent to 40\% drop in FLOPs and parameters. ViT-Slim outperforms all existing methods with substantially fewer FLOPs and parameters and increases accuracy by $\sim$0.6\% of the pre-trained DeiT-B. Note that although our accuracy is comparable to AutoFormer-B, our searching resource is only {\bf 1/10} of it with smaller model size and FLOPs.
	\begin{table}[]
\centering
\resizebox{0.48\textwidth}{!}{
\setlength\tabcolsep{1.5pt}
\begin{tabular}{@{}cc|cc|cc@{}}
Budget & Model & \#Params (M) & FLOPs (B) &  Top-1 (\%) &  Top-5 (\%) \\ \midrule
100 &  DeiT-B      & 86.6       & 17.5       & 81.8         & 95.6          \\
- &  PS-ViT-B \cite{tang2021patch}      & 86.6       & 10.5       & 81.5         & -          \\
- &  S$^2$ViTE-B \cite{chen2021chasing}      & 56.8       & 11.7       & 82.2         & -          \\
- &  GLiT-B \cite{chen2021glit}      & 96.1       & 17.0       & 82.3         & -          \\
- &  AutoFormer-B \cite{AutoFormer}      & 54.0       & 11.0       & \textbf{\uline{82.4}}         & 95.7         \\
60  &  ViT-Slim-B      & \textbf{\uline{52.6}}       & \textbf{\uline{10.6}}       &   \textbf{\uline{82.4}}       & \textbf{\uline{96.1}}\\ \midrule
100        &   \multirow{3}{*}{Swin-T}      & 28.3       & 4.5       & 81.3         & 95.5          \\
80         &          & 22.3       & 3.8       & 81.3         & 95.5          \\
70         &          & 19.4       & 3.4       & 80.7         & 95.4
\end{tabular}
}
\vspace{-0.1in}
\caption{Performance of {\bf DeiT-B}~\cite{pmlr-v139-touvron21a} ({\bf ViT-Slim-B$_{\bf PS}$}) and {\bf Swin-T}~\cite{liu2021swin} for MLP and MHSA joint dimension search. Budget indicates the \% of active MHSA and MLP dimensions across the network respectively.}
\label{tab:baseswin}
\vspace{-0.14in}
\end{table}
	
	We also perform a search with the same hyperparameters as ViT-Slim$_{\bf{PS}}$ on Swin-T \cite{liu2021swin} and present the results in Table \ref{tab:baseswin}. The models are retrained with the same policy as in \cite{liu2021swin}. Final accuracy is intact at 80\% budget but drops at 70\% budget. This is partly because of the fact that Swin is a carefully designed hierarchical architecture which already maximises dimensions at every layer and partly because we don't do a thorough hyperparameter search for sparsity weights specifically for Swin-T. But, still we achieve a fair amount of compression while keeping the accuracy intact.
	
	\vspace{-0.06in}
	\subsection{Comparison to State-of-the-art Approaches}
	\vspace{-0.04in}
	
	\begin{table}[]
\vspace{-0.08in}
\centering
\resizebox{0.48\textwidth}{!}{
\setlength\tabcolsep{2pt}
\begin{tabular}{@{}c|cc|cc@{}}
Model          & \#Params (M) & FLOPs (B) & Top-1 (\%) & Top-5 (\%) \\ \midrule
DeiT - S \cite{pmlr-v139-touvron21a}       & 22.0       & 4.6       & 79.9       & 95.0       \\
GLiT - S \cite{chen2021glit}      & 24.6       & 4.4       & 80.5       & -           \\
DynamicViT - S \cite{rao2021dynamicvit} & 22.0       & 4.0       & 79.8       & -           \\
\bf{ViT-Slim$_{\bf{PS}}$}       & \textbf{\uline{17.7}}       & \textbf{\uline{3.7}}       & \textbf{\uline{80.6}}       & \textbf{\uline{95.3}}       \\\midrule
S$^2$ViTE - S \cite{chen2021chasing}     & 14.6       & 3.1       & 79.2       & -           \\
DynamicViT - S \cite{rao2021dynamicvit} & 22.0       & 2.9       & 79.3       & -           \\
PS-ViT-S \cite{tang2021patch} & 22.0       & 2.7       & 79.4       & -           \\
ViTAS - E \cite{su2021vision}     & 12.6       & 2.7       & 77.4       & 93.8       \\ 
S$^2$ViTE+ - S \cite{chen2021chasing}    & 14.6          & 2.7       & 78.2       & -           \\
\bf{ViT-Slim$_{\bf{JS}}$}      & {15.7}      & \textbf{\uline{3.1}}       & \textbf{\uline{79.9}}       & \textbf{\uline{95.0}}       \\ 
\bf{ViT-Slim$_{\bf{JS}}$}      & 15.7       & \textbf{\uline{2.8}}       & \textbf{\uline{79.5}}       & \textbf{\uline{94.6}}
\end{tabular}
}
\vspace{-0.1in}
\caption{Comparison with SOTA ViT search and compression methods on ImageNet-1K. ``$\bf{PS}$'' is the partial search or parameter search, i.e., MHSA+MLP dimensions, since {\em Patch Selection} will not affect the number of parameters but only help reduce FLOPs in our models. ``$\bf{JS}$'' is our final joint-dimension search.}
\label{tab:sota}
\vspace{-0.10in}
\end{table}
	We compare both our model families: ViT-Slim$_{\bf{PS}}$ and ViT-Slim$_{\bf{JS}}$ with existing efficient transformer architecture search and compression methods as shown in Table \ref{tab:sota}. Our method outperforms all of them at different target parameters and FLOPs. GLiT-S \cite{chen2021glit} improves accuracy (80.5\%) over baseline DeiT-S but with an additional parameter increase and minimal FLOPs reduction. Our ViT-Slim$_{\bf{PS}}$ achieves better accuracy (80.6\%) and substantially decreases both FLOPs and Parameters at the same time. Our Vit-Slim$_{\bf{JS}}$ model allows for up to 30\% parameter reduction and greater than 30\% FLOPs reduction while matching the performance of it's supernet (DeiT-S). Even further reducing FLOPs doesn't cause a drastic degradation in performance of ViT-Slim$_{\bf{JS}}$ in contrast to others. PS-ViT-S \cite{tang2021patch} and Dynamic-ViT-S \cite{rao2021dynamicvit} performs comparably to our method with respect to FLOPs, but they do not bring in any improvement in parameter efficiency.
	
	\vspace{-0.03in}
	\subsection{Transfer Learning on Downstream Datasets}
	\vspace{-0.02in}
	\begin{table}[]
\centering
\resizebox{0.48\textwidth}{!}{
\begin{tabular}{c|cc|cccc}
\multicolumn{1}{c|}{Model} & \begin{tabular}[c]{@{}c@{}} \#Params \\  (M)\end{tabular} & \begin{tabular}[c]{@{}c@{}} FLOPs \\  (B)\end{tabular} & \multicolumn{1}{c}{C100} & \multicolumn{1}{c}{C10} & \multicolumn{1}{c}{iNat-19} & iNat-18 \\ \midrule
DEIT-S                     & 22.0       & 4.6       & 87.80      & 98.56    & 75.35     & 69.02         \\
\bf{ViT-Slim$_{\bf{PS}}$}                 & \bf15.6       & \bf3.3       &  \textbf{\uline{88.16}}      &  \textbf{\uline{98.70}}    &  \textbf{\uline{76.67}}     &  \textbf{\uline{69.83}}         \\      
\end{tabular}
}
\vspace{-0.1in}
\caption{Transfer learning accuracy of {\bf DeiT-S}~\cite{pmlr-v139-touvron21a} and {\bf ViT-Slim$_{\bf PS}$} on CIFAR-100 (C100), CIFAR-10 (C10) \cite{CIFAR}, iNaturalist-2018 (iNat-18) and iNaturalist-2019 \cite{van2018inaturalist} (iNat-19) datasets. All models are searched and pre-trained on ImageNet-1K.}.
\label{tab:transfer}
\vspace{-0.19in}
\end{table}
	We analyse the performance of our searched and retrained model on various downstream classification tasks. We provide results on CIFAR-10, CIFAR-100 \cite{CIFAR}, inaturalist-2018 \cite{van2018inaturalist} and inaturalist-2019 datasets as shown in Table \ref{tab:transfer}. ViT-Slim$_{\bf{PS}}$ is retrained at 70\% budget and they consistently outperform DeiT-S baseline across datasets. An important point to note is that ViT-Slim architectures were searched on ImageNet-1K \cite{deng2009imagenet} and not on respective downstream datasets directly, which shows the ability of the same ViT-Slim architectures to transfer well on other downstream tasks too.
	
	\vspace{-0.01in}
	\section{Visualization and Analysis}
	
	\vspace{-0.03in}
	\subsection{Searched Architecture}
	\vspace{-0.02in}
	
	\begin{figure*}[t]
		\centering
		\includegraphics[width=14.5cm]{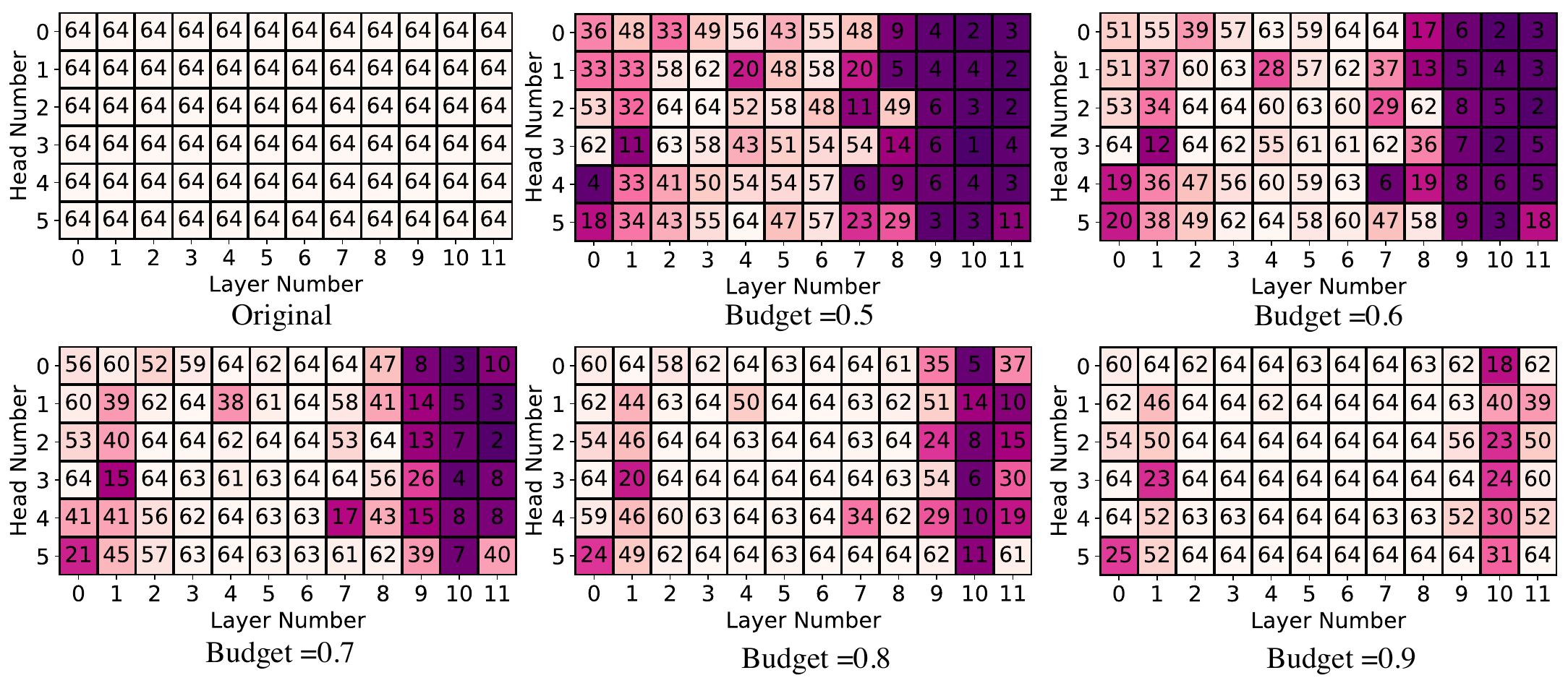}
		\vspace{-0.17in}
		\caption{Layerwise Attention Head Dimensions at various budgets (50\%, 60\%, 70\%, 80\% and 90\%) for {\bf ViT-Slim} model.}
		\label{fig:attnhead}
		\vspace{-0.16in}
	\end{figure*}
	
	\noindent{\textbf{Layerwise attention head dimensions at various budgets.}}
	Figure \ref{fig:attnhead} shows the MHSA modules across DeiT-S searched model. There are a total of 12 MHSA modules each with 6 attention heads. Numbers inside the grids indicate the dimension size of that particular head. It can be seen that at low budgets, deeper layers have the least dimension sizes. Most of the dimensions are intact in the middle of the network and a moderate reduction in dimensions is done in the beginning of the network. Self-attention mechanism is required in the middle and to some extent at the beginning of the network when the patches are distinct and information exchange between them is required.
	
	\begin{figure}[t]
		\centering
		\includegraphics[width=6.4cm]{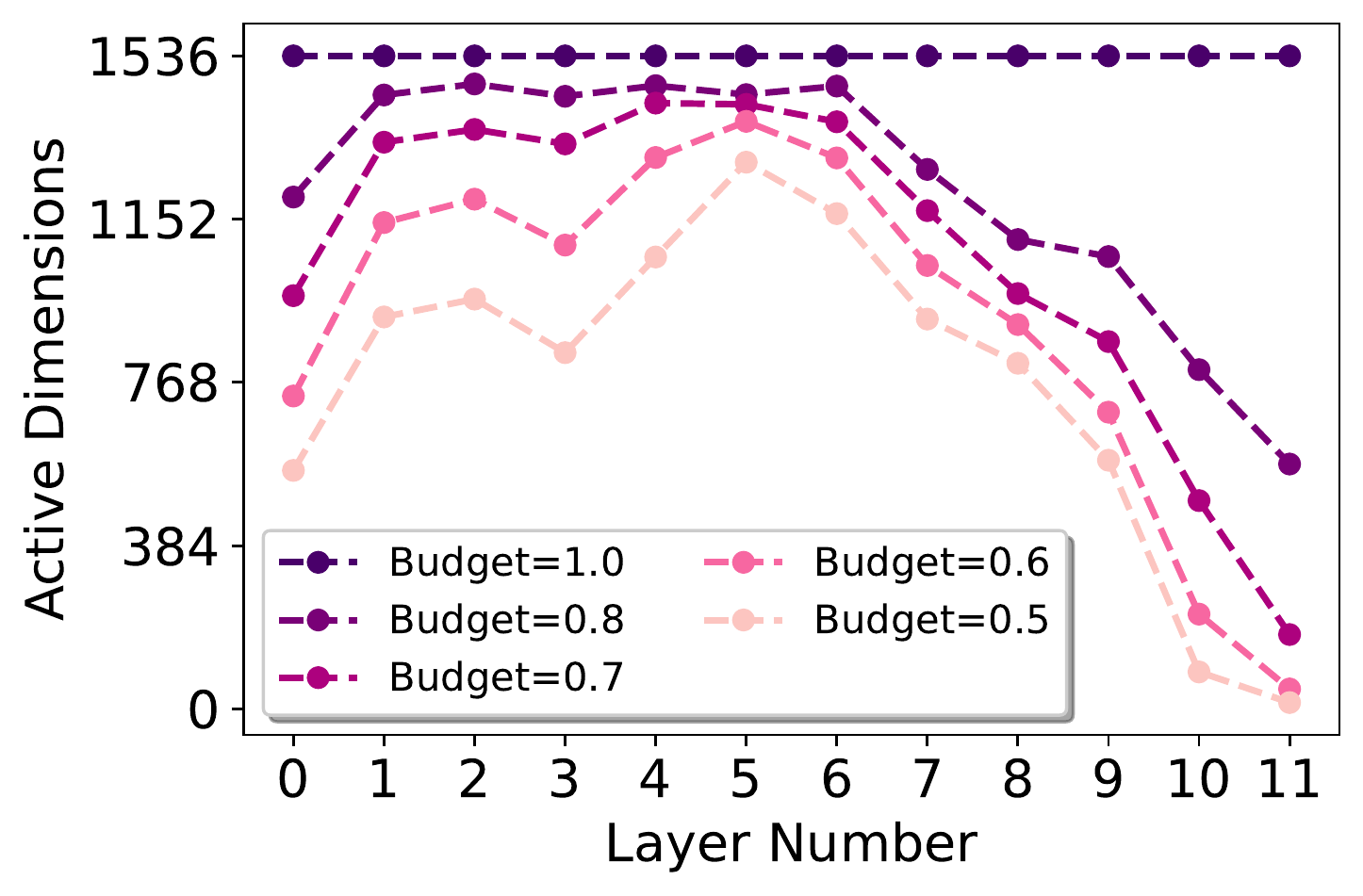}
		\vspace{-0.15in}
		\caption{Layerwise MLP dimensions at various budgets (50\%, 60\%, 70\% and 80\%) for {\bf ViT-Slim} model.}
		\vspace{-0.11in}
		\label{fig:mlp}
	\end{figure}
	
	\noindent{\textbf{Layerwise MLP dimensions at various budgets.}}
	Similarly, Figure \ref{fig:mlp} shows the MLP module dimensions across DeiT-S searched model at various budgets. The pattern across layers is similar to that of MHSA, where deeper layers have a greater degree of reduced dimensions as compared to the earlier layers. The deeper layers have maximum dimensions removed, while the middle layers have maximum dimensions intact. This is again in agreement with the fact that the majority of features are already learnt in the earlier layers, enabling deeper layers to consist of smaller dimension sizes.

	\begin{figure}[t]
		\centering
		\begin{tabular}{c}
			\includegraphics[width=8.1cm]{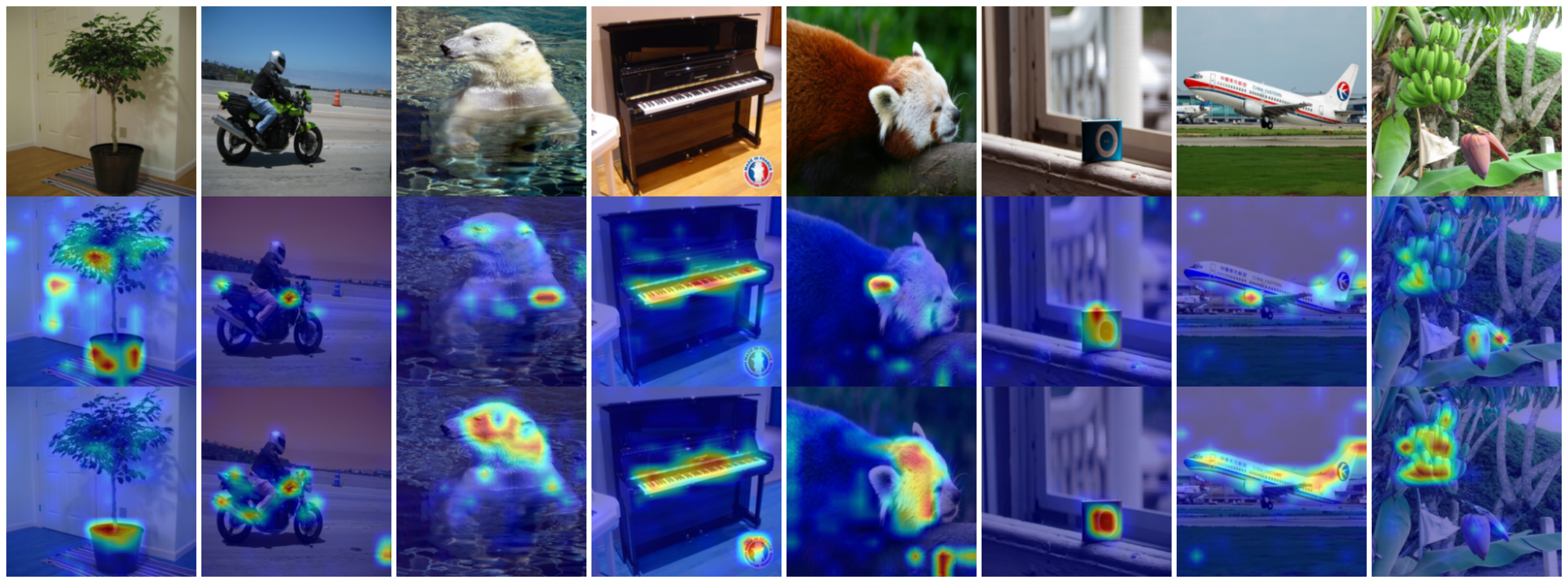}\\
			\includegraphics[width=8.1cm]{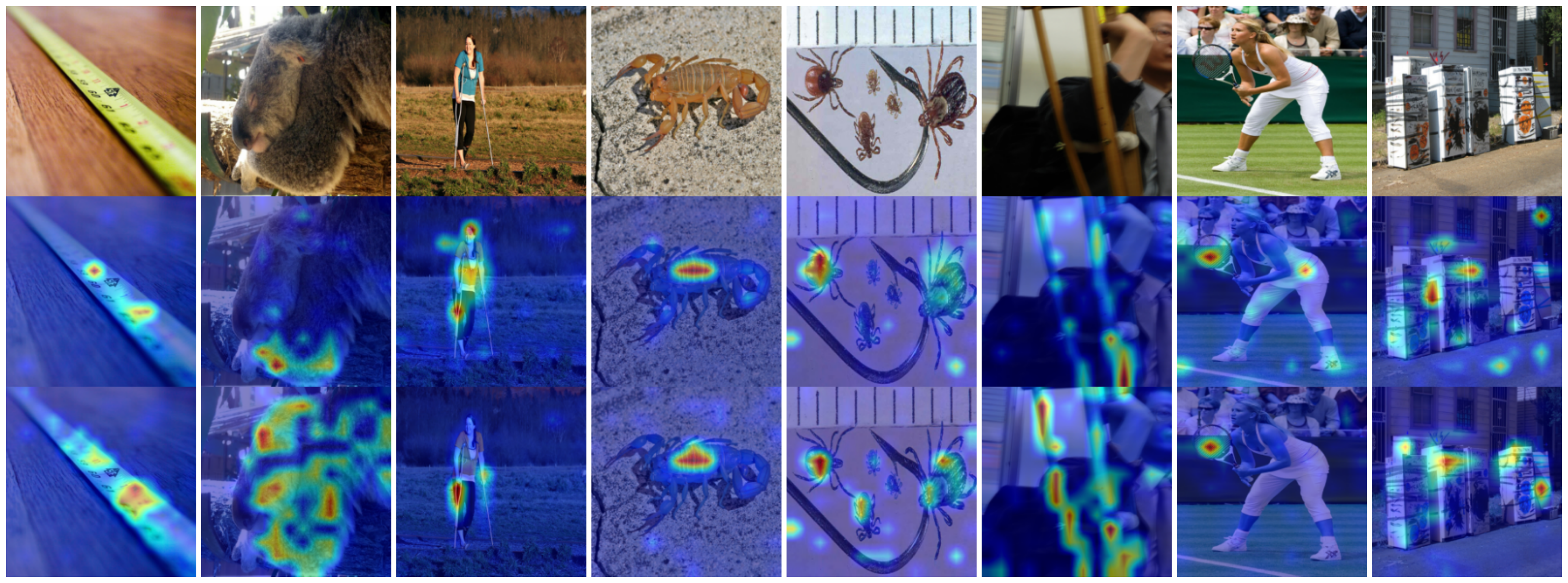}
		\end{tabular}
		\vspace{-0.16in}
		\caption{Class wise visualisation of ImageNet-1K images with method presented in \cite{Chefer_2021_CVPR}. First row are the original images, second row represent visualisations from {\bf DeiT-S} and third row represents {\bf ViT-Slim$_{\bf{JS}}$} at 70\% budget.}
		\label{fig:attnmap}
		\vspace{-0.17in}
	\end{figure}
	
	\vspace{-0.06in}
	\subsection{{Attention Maps Visualization} }
	\vspace{-0.03in}
	We adopt the method presented in \cite{Chefer_2021_CVPR} which employs Deep Taylor Decomposition to calculate local relevance and then propagates these relevancy scores through the layers to generate a final relevancy map. Class-wise visualisation of randomly chosen ImageNet-1K images is shown in Figure \ref{fig:attnmap}. ViT-Slim$_{\bf{JS}}$ focuses better on class-specific important areas as compared to DeiT-S and thus achieves better performance. This also shows that ViT-Slim$_{\bf{JS}}$ has better interpretability and thus helps in domains where interpretability and explainability of deep networks are important. 
	
	\vspace{-0.06in}
	\section{Conclusion}
	\vspace{-0.03in}
	
	We have presented {\em ViT-Slim}, a flexible and efficient searching strategy for subnet discovery on vision transformers leveraging the model sparsity. The proposed method can jointly search on all three dimensions in a ViT, including: layerwise tokens/patches, MHSA and MLP modules end-to-end. We identified that the global importance factors are crucial, and designed additional differentiable soft masks on different modules to reflect the individual importance of dimension. Moreover, the $\ell_1$-sparsity is imposed to force the masks to be sparse during searching. Extensive experiments are conducted on ImageNet-1K and downstream datasets using a variety of ViT architectures to demonstrate the efficiency and effectiveness of our proposed method.
	
	{\small
		\bibliographystyle{ieee_fullname}
		\bibliography{egbib}
	}
	
\end{document}